\documentclass[11pt]{article}

\usepackage[final]{acl}

\usepackage{times}
\usepackage{latexsym}

\usepackage[T1]{fontenc}

\usepackage[utf8]{inputenc}

\usepackage{microtype}

\usepackage{inconsolata}

\usepackage{graphicx}

\usepackage{booktabs}
\usepackage{multirow}
\usepackage{multicol}
\usepackage{amssymb}
\usepackage{amsmath}
\usepackage{amsfonts}
\usepackage{tabularx}
\title{VideoCuRL: Video Curriculum Reinforcement Learning with Orthogonal Difficulty Decomposition}

\author{
 \textbf{Hongbo Jin}\quad 
 \textbf{Kuanwei Lin} \quad
  \textbf{Wenhao Zhang} \quad
  \textbf{Yichen Jin} \quad
 \textbf{Ge Li\thanks{corresponding author}}
\\
 School of Electronic and Computer Engineering, \\
 Peking University,
\\
 \small{
   \textbf{Correspondence:} \href{mailto:hbjin25@stu.pku.edu.cn}{hbjin25@stu.pku.edu.cn}
 }
}

\begin{document}
\maketitle

\begin{abstract}
Reinforcement Learning (RL) is crucial for empowering Video-LLMs with complex spatiotemporal reasoning. However, current RL paradigms predominantly rely on random data shuffling or naive curriculum strategies based on scalar difficulty metrics. We argue that scalar metrics fail to disentangle two orthogonal challenges in video understanding: Visual-Temporal Perception Load and Cognitive Reasoning Depth. To address this, we propose VideoCuRL, a novel framework that decomposes difficulty into these two axes. We employ efficient, training-free proxies—optical flow/keyframe entropy for visual complexity and Calibrated Surprisal for cognitive complexity—to map data onto a 2D curriculum grid. A competence-aware Diagonal Wavefront strategy then schedules training from base alignment to complex reasoning. Furthermore, we introduce Dynamic Sparse KL and Structured Revisiting to stabilize training against reward collapse and catastrophic forgetting. Extensive experiments show that VideoCuRL surpasses strong RL baselines on reasoning (+2.5\% on VSI-Bench) and perception (+2.9\% on VideoMME) tasks. Notably, VideoCuRL eliminates the prohibitive inference overhead of generation-based curricula, offering a scalable solution for robust video post-training.

\end{abstract}

\section{Introduction}
Large Multimodal Models (LMMs) have catalyzed significant advancements in video understanding, enabling systems capable of following complex instructions, aligning vision and language, and reasoning over extended temporal contexts ~\cite{li2025videochat,li2024llava,bai2025qwen2}.
By coupling powerful visual encoders with Large Language Models (LLMs), modern Video-LLMs achieve remarkable performance on benchmarks~\cite{xu2017video,yu2019activitynet,lei2018tvqa,jang2017tgif} spanning action recognition, temporal grounding, and video question answering.
Nevertheless, robust spatiotemporal video reasoning remains an open challenge.
Current models often falter on tasks that demand long-range temporal integration, causal inference, or excessive perception load.

Reinforcement Learning (RL)~\cite{schulman2017proximal,ouyang2022training,shao2024deepseekmath} has emerged as a promising post-training paradigm to address this gap.
In contrast to Supervised Fine-Tuning (SFT), RL incentivizes exploration and has demonstrated clear gains on complex video reasoning tasks~\cite{feng2025video}.
However, current video reinforcement learning methods lack sophisticated curriculum design and mostly rely on random data shuffling.
This design implicitly assumes that all samples are equally suitable throughout training.
In practice, this assumption is brittle for video: training on samples misaligned with the model’s evolving competence—either too trivial or too difficult—results in inefficient optimization, unstable gradients, and catastrophic forgetting.

Curriculum Learning (CL)~\cite{narvekar2020curriculum,wang2021survey,soviany2022curriculum} offers an alternative by organizing training data from easy to hard. Yet, curriculum strategies tailored for video understanding remain underdeveloped.
Existing methods mainly rely on simple scalar difficulty metrics, such as sequence length, model perplexity, or response accuracy~\cite{li2025adacurl,dong2025videotg}.
We argue that such scalar metrics are insufficient for video domain,
where difficulty is inherently multi-dimensional.
A clip may impose heavy visual-temporal perception load while requiring only shallow reasoning, or conversely, be visually simple yet demand deep causal reasoning.
\begin{figure}[h]
    \centering
    \includegraphics[width=\linewidth]{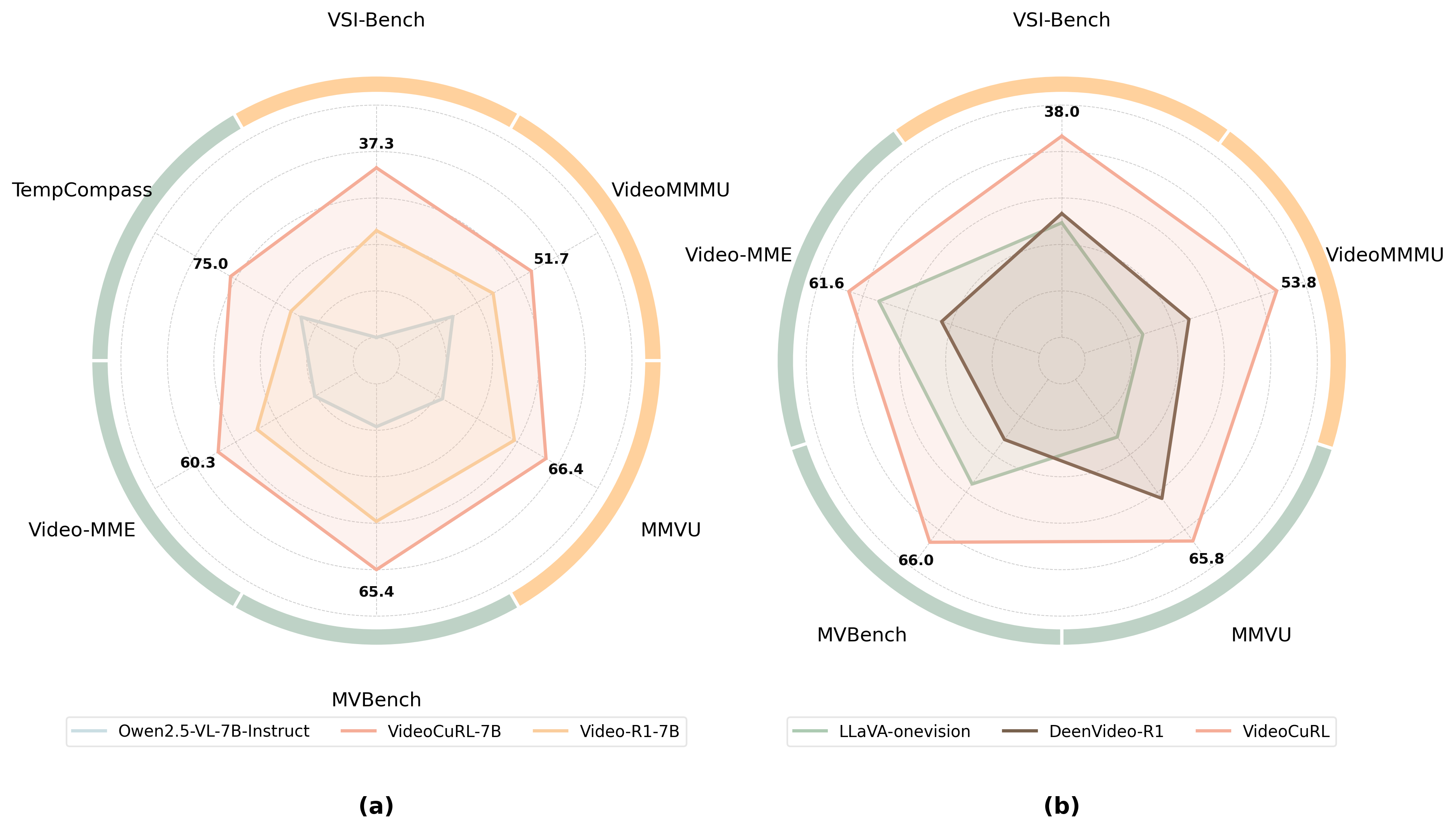}
    \caption{Multidimensional performance comparison.
    (Left) Comparison with baselines.
    (Right) Comparison with other open-source models.}
    \label{fig:radar}
\end{figure}
Collapsing these orthogonal challenges into a single scalar obscures the distinction between perceptual and cognitive complexity,
inevitably conflating heterogeneous learning demands and resulting in unstable training trajectories.

To address this,
we introduce \textbf{VideoCuRL},
a novel curriculum reinforcement learning framework that explicitly disentangles video difficulty into two orthogonal dimensions:
Visual-Temporal Perception Load and Cognitive Reasoning Depth.
Instead of relying on expensive model-based generation, VideoCuRL utilizes efficient, training-free proxies to map data onto a 2D curriculum grid.
This allows us to employ a competence-aware Diagonal Wavefront strategy, effectively scheduling training from base alignment to complex reasoning while avoiding the prohibitive overhead of existing methods.
To guarantee optimization stability, we further integrate Dynamic Sparse KL and Structured Revisiting. These mechanisms counteract the twin challenges of reward sparsity in complex reasoning and catastrophic forgetting of perceptual skills, effectively preventing policy degradation while ensuring the retention of fundamental spatiotemporal capabilities.
We evaluate VideoCuRL on a comprehensive suite of benchmarks, spanning complex video reasoning~\cite{yang2025thinking,zhao2025mmvu,hu2025video} and general video understanding~\cite{fu2025video,li2024mvbench,liu2024tempcompass}.
Empirical results (see Figure \ref{fig:radar}) demonstrate that VideoCuRL consistently outperforms strong RL baselines and prior curriculum strategies, achieving significant gains on reasoning-intensive tasks while simultaneously improving perceptual robustness. Notably, VideoCuRL matches or exceeds the performance of generation-based curricula with negligible computational overhead, confirming that our orthogonal difficulty modeling is both effective and scalable.

In summary, our contributions are threefold:
\begin{itemize}
    \item We propose Orthogonal Difficulty Decomposition, a novel paradigm that challenges traditional scalar metrics by explicitly disentangling video difficulty into Visual-Temporal and Cognitive dimensions.
    \item We introduce VideoCuRL, a scalable two-dimensional curriculum reinforcement learning framework with a competence-aware Diagonal Wavefront scheduler and robust optimization mechanisms.
    \item Extensive experiments show that VideoCuRL substantially outperforms existing video RL baselines, 
    securing clear gains on reasoning-intensive tasks while preserving fundamental visual capabilities.
    It validates the superiority of our 2D curriculum over random shuffling and scalar-based strategies.
\end{itemize}

\section{Related Work}

\subsection{Large Multimodal Models for Video Understanding}
Video understanding has been significantly advanced by Large Multimodal Models (LMMs)~\cite{lin2024video,zhang2024video,weng2024longvlm,zhang2025videollama}.
Early approaches adapted image-based architectures via pooling mechanisms~\cite{xu2024pllava,liu2024ppllava} or temporal adapters~\cite{liu2024bt} for basic modal alignment.
Current state-of-the-art Video-LLMs~\cite{bai2025qwen2,wang2025internvl3} integrate visual-temporal encoders with Large Language Models (LLMs) to handle extended temporal contexts.
Despite these improvements,
a critical disconnect persists between visual perception and cognitive reasoning.
Standard Supervised Fine-Tuning (SFT) often fails to elicit robust spatiotemporal reasoning,
as models tend to exploit language priors or shallow visual cues rather than grounding logic in temporal dynamics.
Consequently, SFT is increasingly seen as insufficient for mastering complex video tasks. 

\begin{figure*}[h]
    \centering
    \includegraphics[width=\linewidth]{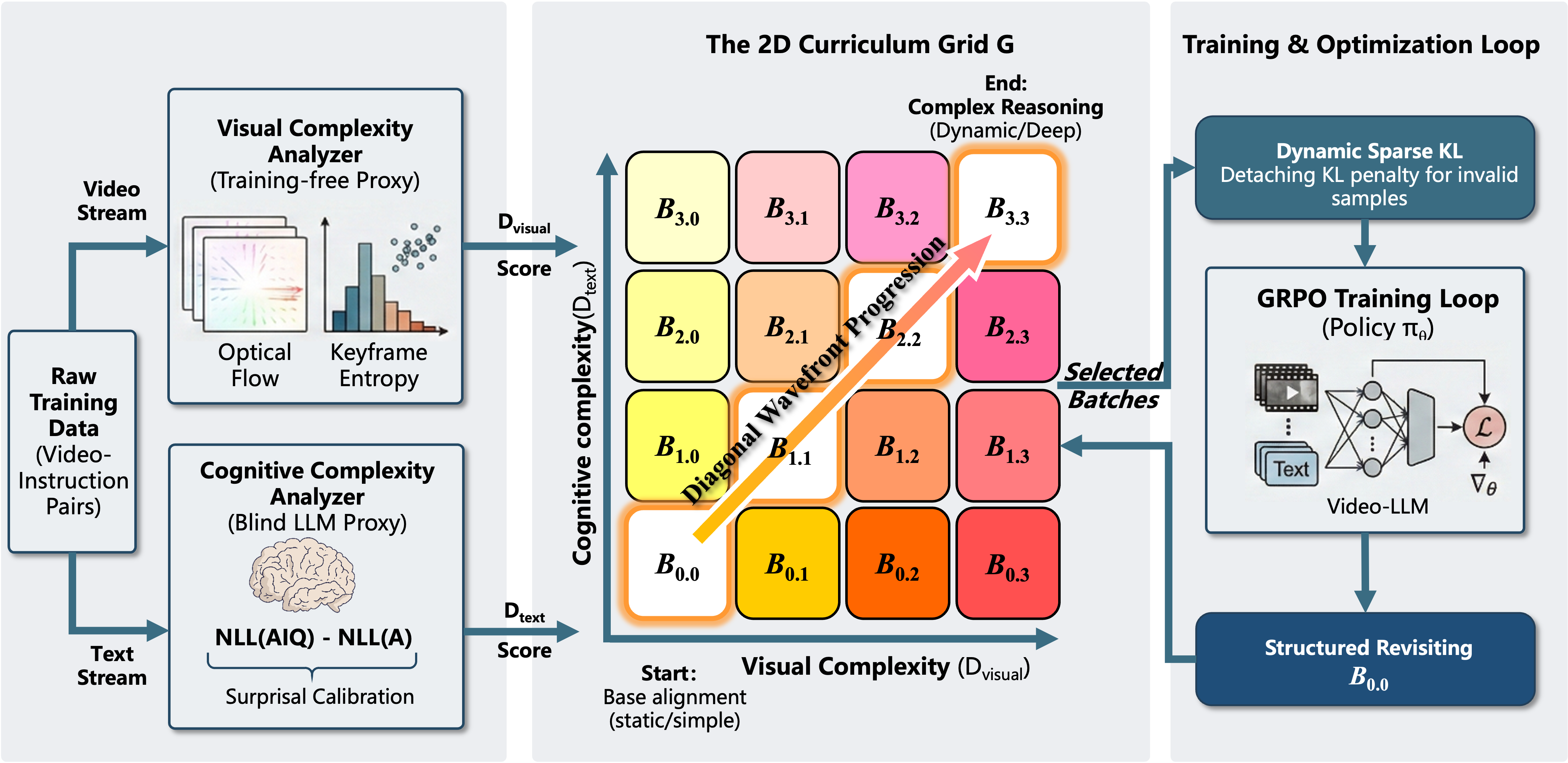}
    \caption{\textbf{Overview of VideoCuRL}. VideoCuRL disentangles video difficulty into orthogonal visual-temporal ($D_{visual}$) and cognitive ($D_{text}$) dimensions.
    A competence-aware Diagonal Wavefront strategy schedules training across a 2D grid, supported by Dynamic Sparse KL and Structured Revisiting to stabilize RL and prevent forgetting.}
    \label{fig:overall pipeline}
\end{figure*}

\subsection{Reinforcement Learning in Multimodal Post-training}
Reinforcement Learning (RL) has become a standard paradigm to enhance reasoning and alignment in LLMs~\cite{schulman2017proximal,ouyang2022training,rafailov2023direct,shao2024deepseekmath} and has recently extended to multimodal domains~\cite{yu2024rlhf,wang2024mdpo,huang2025vision,shen2025vlm}.
By optimizing task-level rewards, RL enables exploration beyond supervised signals.
However, its application to video understanding remains challenging.
Most existing methods~\cite{feng2025video,li2025videochatr1,zhang2025tinyllava} rely on random data sampling, implicitly treating all samples as equally suitable throughout training.
This assumption is inefficient for video where samples vary dramatically in both visual perception load and cognitive reasoning demand.
To address this,
we propose VideoCuRL, the first curriculum reinforcement learning framework that explicitly decouples video difficulty into two orthogonal dimensions: visual and cognitive.

\section{Methodology}
\subsection{Overview}
\begin{figure*}[h]
    \centering    \includegraphics[width=\linewidth]{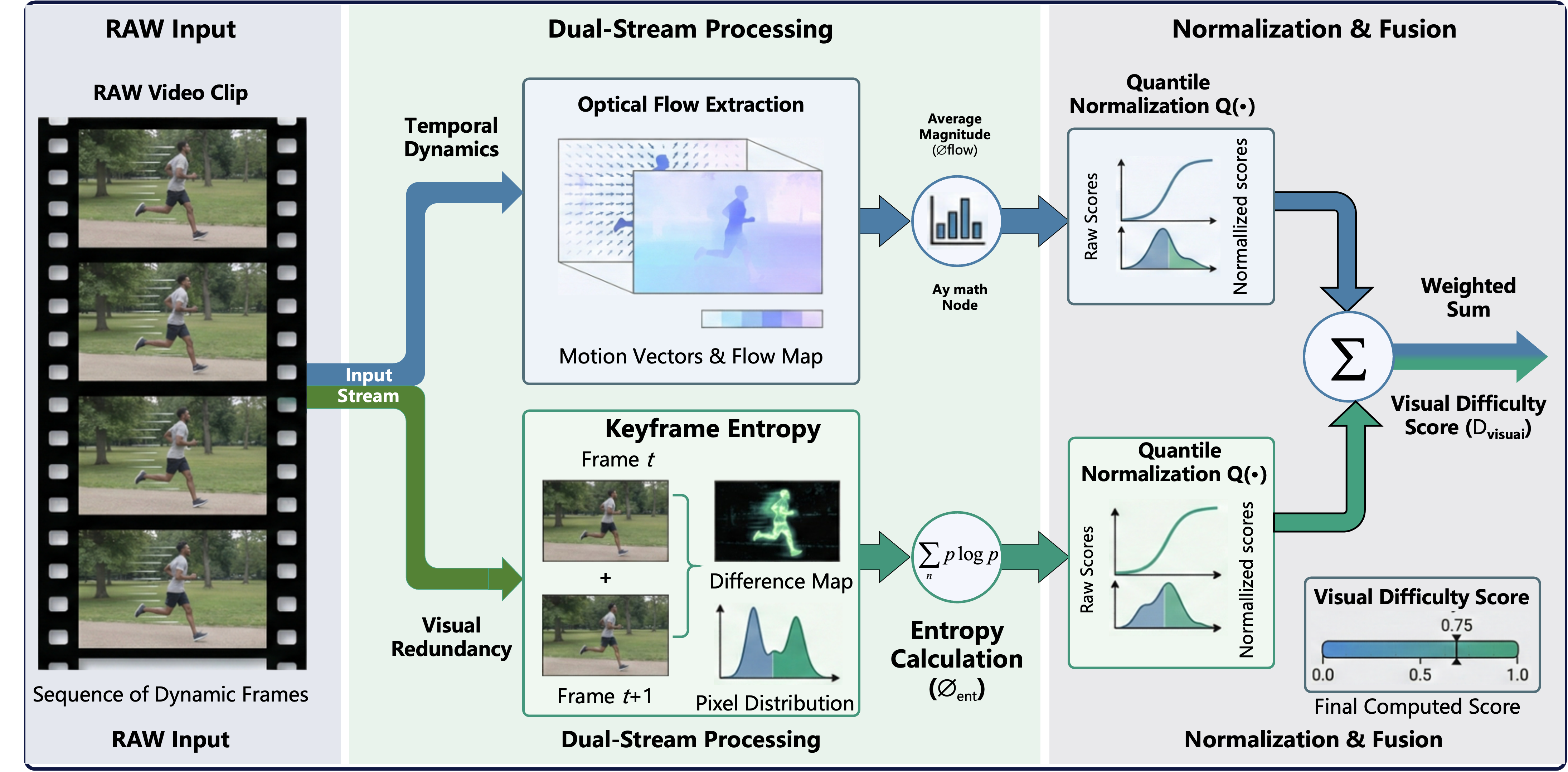}
    \caption{\textbf{Visual Complexity Estimation}.
    Perceptual load is quantified via dual-stream processing:
    Motion Intensity ($\phi_{flow}$) from optical flow and Information Density ($\phi_{ent}$) from frame-difference entropy.
    Scores are fused and quantile-normalized to ensure a balanced difficulty distribution.}
    \label{fig:visual complexity}
\end{figure*}

Applying Reinforcement Learning (RL) combined with Curriculum Learning (CL) to Video-LLMs remains unexplored.
Current video post-training paradigms predominately rely on Supervised Fine-Tuning (SFT) or RL with random data shuffling, neglecting the progressive nature of human learning.
Therefore we propose VideoCuRL, a framework that disentangles difficulty into two orthogonal axes: Visual-Temporal Complexity ($D_{visual}$) and Cognitive Reasoning Complexity ($D_{text}$).
We construct a 2D curriculum grid and employ a diagonal wavefront scheduling strategy equipped with structured revisiting and adaptive optimization to robustly enhance reasoning capabilities.

\subsection{Visual-Temporal Complexity ($D_{visual}$)}

Standard video instruction tuning treats all video clips equally, ignoring that a static shot and a fast-paced montage impose vastly different perceptual loads.
We quantify perception load by synthesizing motion dynamics and information redundancy:

\textbf{Motion Intensity ($\phi_{flow}$)}:
To quantitatively measure temporal dynamics,
we utilize dense optical flow to capture pixel-wise displacement between consecutive frames.
Let a video sequence be denoted as $V = \{F_1, F_2, \dots, F_T\}$, where $F_t \in \mathbb{R}^{H \times W \times 3}$.
For each pair of adjacent frames $(F_t, F_{t+1})$, we compute the dense optical flow field $\mathbf{V}_t \in \mathbb{R}^{H \times W \times 2}$,
where each coordinate $(x, y)$ contains a displacement vector $\mathbf{v}_{t,x,y} = (u_{t,x,y}, v_{t,x,y})$ representing the horizontal and vertical shift.
We define the frame-level motion score $m_t$ as the spatial average of Euclidean norms of these vectors:
\begin{equation}
\begin{split}
m_t &= \frac{1}{H \times W} \sum_{x=1}^{H} \sum_{y=1}^{W} \|\mathbf{v}_{t,x,y}\|_2 \\
    &= \frac{1}{H \times W} \sum_{x,y} 
        \sqrt{u_{t,x,y}^2 + v_{t,x,y}^2}
\label{eq:frame_motion_score}  
\end{split}
\end{equation}
The video-level Motion Intensity $\phi_{flow}$ is obtained by temporally pooling the frame-level scores, representing the global kinetic energy of the clip:
\begin{equation}
    \phi_{flow}(V) = \frac{1}{T-1} \sum_{t=1}^{T-1} m_t
    \label{eq:video_motion_score}
\end{equation}
High $\phi_{flow}$ values indicate rapid object movement or significant camera shifts, necessitating robust temporal tracking capabilities from the model.

\textbf{Visual Information Density ($\phi_{ent}$)}:
While optical flow captures kinetic magnitude, it does not account for visual redundancy.
To quantify the diversity of visual information, we calculate the entropy of pixel-wise differences between consecutive frames.
We specifically employ temporal difference entropy rather than spatial frame entropy. This design choice effectively disentangles visual texture complexity (handled well by the pre-trained spatial encoder) from temporal information density (the primary challenge for video reasoning), preventing static scenes with high-frequency textures from being misclassified as difficult samples.

Let $\Delta F_t(x,y)$ be the grayscale intensity of the difference map at coordinates $(x,y)$, taking values in the integer range $[0, 255]$:
\begin{equation}
    \Delta F_t = | \text{Gray}(F_{t+1}) - \text{Gray}(F_t) |
    \label{eq:gray scale}
\end{equation}
We first compute the normalized histogram (probability distribution) $p_t$ of these intensity values.
For each intensity level $k \in \{0, \dots, 255\}$, the probability $p_t(k)$ is given by:
\begin{equation}
    p_t(k) = \frac{\text{Count}(\Delta F_t(x,y) = k)}{H \times W}
    \label{eq:probability}
\end{equation}
where $\sum_{k=0}^{K-1} p_t(k) = 1$.
The visual information entropy $E_t$ for the frame difference is then calculated as:
\begin{equation}
    E_t = - \sum_{k=0}^{K-1} p_t(k) \log_2 (p_t(k) + \epsilon)
    \label{eq:entropy}
\end{equation}
(Note: We add a small constant $\epsilon$ simply to ensure numerical stability where $p_t(k)=0$.)

A low $E_t$ implies that most pixels have similar changes (e.g., a static background or uniform motion), indicating high redundancy.
Conversely, a high $E_t$ suggests complex, non-uniform visual changes. The final video-level density score $\phi_{ent}$ is the temporal average of these entropy values:
\begin{equation}
    \phi_{ent}(V) = \frac{1}{T-1} \sum_{t=1}^{T-1} E_t
    \label{eq:temporal entropy}
\end{equation}
This metric effectively filters out visually monotonous clips from information-dense footage, complementing the motion intensity score.

\textbf{Final Difficulty ($D_{visual}$)}:
The final visual difficulty score is a weighted combination, normalized via quantile normalization $\mathcal{Q}$ to ensure a uniform distribution across the $[0, 1]$ interval:
\begin{equation}
\begin{split}
D_{\text{visual}}(i) &= \alpha \cdot \mathcal{Q}\bigl(\phi_{\text{flow}}(v_i)\bigr) \\
&\quad + (1-\alpha) \cdot \mathcal{Q}\bigl(\phi_{\text{ent}}(v_i)\bigr)
\end{split}
\label{eq:final difficulty}
\end{equation}

\subsection{Cognitive Reasoning Complexity ($D_{text}$)}

\textbf{The Bias of Raw Perplexity.}
A naive approach for estimating cognitive difficulty is to use the raw perplexity (or Negative Log-Likelihood, NLL) of the ground-truth answer $A$ given the question $Q$. 
However, this metric is biased by the \textit{language prior}: answers containing low-frequency vocabulary yield high NLL scores, even if the reasoning is trivial.
This leads to a misalignment where "linguistically rare" is mistaken for "logically hard."

\textbf{Calibrated Surprisal}.
To mitigate this, we measure the Information Gain provided by the question $Q$ toward the answer $A$.
We define the cognitive complexity score $S_{cog}$ as the reduction in uncertainty, which effectively isolates the reasoning requirement from inherent linguistic unpredictability. Formally, for an answer sequence $A = \{w_1, w_2, \dots, w_n\}$, we calculate the difference between the conditional and unconditional NLL using the chain rule of probability:
\begin{equation}
    \begin{split}
        S_{cog}(A, Q) = \sum_{t=1}^{|A|} \underbrace{-\log \pi_{\phi}(w_t | w_{<t}, Q)}_{\text{Conditional Surprisal}} \\ - \sum_{t=1}^{|A|} \underbrace{(-\log \pi_{\phi}(w_t | w_{<t}))}_{\text{Language Prior}}
    \end{split}
    \label{eq:Calibrated Surprisal}
\end{equation}
A high $S_{cog}$ indicates that the answer $A$ is not easily predictable from the question $Q$ alone without strong reasoning,
whereas a low score implies high linguistic predictability.

Crucially, this metric relies solely on a "Blind" LLM processing text pairs, avoiding the massive overhead of video encoding or autoregressive video decoding.
Finally,
we map the raw scores to a uniform distribution via Quantile Normalization to obtain the final cognitive difficulty dimension:
\begin{equation}
    D_{text}(i) = \mathcal{Q}(S_{cog}(A_i, Q_i))
    \label{eq:normalize}
\end{equation}
By combining $D_{text}$ and $D_{visual}$, we construct the orthogonal basis for our curriculum grid $\mathcal{G}$.

\subsection{Diagonal Wavefront Scheduling}
To effectively navigate the disjoint difficulty landscape,
we map the training dataset $\mathcal{D}$ into a $K \times K$ grid $\mathcal{G}$.
Each bucket $B_{i,j}$ encapsulates samples with visual complexity level $i$ and cognitive complexity level $j$.
Instead of a linear progression, which risks overfitting to one modality,
we propose a Diagonal Wavefront strategy driven by model competence.

Competence-Aware Gating:
We define the Local Competence Score ($LCS_{i,j}$) as the exponential moving average of rewards obtained from samples in bucket $B_{i,j}$ at step $t$:
\begin{equation}
  LCS_{i,j}^{(t)} = \gamma \cdot LCS_{i,j}^{(t-1)} + (1 - \gamma) \cdot \bar{R}_{batch}  
  \label{eq:LCS}
\end{equation}
where $\bar{R}_{batch}$ is the mean reward of the current batch sampled from $B_{i,j}$.

Wavefront Expansion:
Let $\mathcal{A}_t$ denote the set of active buckets (the "Wavefront").
Training initiates at $B_{0,0}$ (Static Visuals, Atomic Queries). The curriculum unfolds dynamically: a neighboring harder bucket $B_{i', j'}$ (where $i' \ge i, j' \ge j$) is added to $\mathcal{A}_{t+1}$ if and only if the competence in the current frontier exceeds a threshold $\mathcal{T}$:
\begin{equation}
  \text{Expand if } LCS_{i,j}^{(t)} > \mathcal{T} 
  \label{eq:expand}
\end{equation}
This mechanism enforces a "Base Alignment" constraint: the model must demonstrate fundamental proficiency in simpler visual-linguistic grounding before attempting samples with high temporal loads or complex causal chains.
\begin{table*}[h]
\centering
\small
\renewcommand{\arraystretch}{1.1}
\setlength{\tabcolsep}{1.8mm}
\begin{tabularx}{\textwidth}{l c c c c c c c}
\toprule
\multirow{2}{*}{\textbf{Model}} & \multirow{2}{*}{\textbf{Frames}} & \multicolumn{3}{c}{\textbf{Video Reasoning Benchmarks}} & \multicolumn{3}{c}{\textbf{General Video Benchmarks}} \\
\cmidrule(lr){3-5} \cmidrule(lr){6-8}
& & VSI-Bench & VideoMMMU & MMVU & MVBench & TempCompass & VideoMME \\
\midrule

LLaVA-onevision-7B &32&32.4&33.8&49.2&56.7&-&58.2\\
LLaVA-video-7B &32&-&34.4&48.8&58.6&-&63.3\\
VILA-1.5-40B &-&31.2&34.0&-& -&-& 60.1 \\
Kangaroo-8B &-&-&-&-&61.1&62.5&56.0\\
DeepVideo-R1-3B &-&33.0 &40.7&59.0&49.6&63.1&51.1\\
\midrule
\textbf{w/o Curriculum} \\
Qwen2.5-VL-7B-Instruct & 16 &27.7&47.8&59.2&57.4&72.2&53.1  \\
Video-R1-7B & 16 &34.6&49.8&64.2&62.7&72.6&57.4  \\
Qwen2.5-VL-7B-Instruct & 32&30.1&48.1&60.0&59.0&72.6&56.6 \\
Video-R1-7B & 32 & 35.8&52.3&63.8&63.9&73.2&59.3  \\
Qwen2.5-VL-7B-Instruct & 64 & 31.4&50.4&60.0&59.2&72.9&59.6 \\
Video-R1-7B & 64 & 37.1&52.4&63.8&64.8&73.2&61.4\\
\midrule
\textbf{w/ Curriculum} \\
Length-based&16	&35.1	&50.5&	64.5	&63.0	&72.5&	60.2 \\
Generation-based 	&16	&36.5	&51.0	&65.6	&64.7&	73.8&60.4\\
VideoCuRL-7B & 16 & 37.3\textbf{\textcolor{blue}{($\uparrow$2.7)}} &51.7\textbf{\textcolor{blue}{($\uparrow$1.9)}}&66.4\textbf{\textcolor{blue}{($\uparrow$2.2)}}&65.4\textbf{\textcolor{blue}{($\uparrow$2.7)}}&75.0\textbf{\textcolor{blue}{($\uparrow$2.4)}}&60.3\textbf{\textcolor{blue}{($\uparrow$2.9)}}\\
VideoCuRL-7B  & 32 & 38.0\textbf{\textcolor{blue}{($\uparrow$2.2)}}&53.8\textbf{\textcolor{blue}{($\uparrow$1.5)}}&65.8\textbf{\textcolor{blue}{($\uparrow$2.0)}}&66.0\textbf{\textcolor{blue}{($\uparrow$2.1)}}&75.8\textbf{\textcolor{blue}{($\uparrow$2.6)}}&61.6\textbf{\textcolor{blue}{($\uparrow$2.3)}} \\
VideoCuRL-7B  & 64 &39.6\textbf{\textcolor{blue}{($\uparrow$2.5)}}&54.5\textbf{\textcolor{blue}{($\uparrow$2.1)}}&65.7\textbf{\textcolor{blue}{($\uparrow$1.9)}}&66.9\textbf{\textcolor{blue}{($\uparrow$2.1)}}&75.5\textbf{\textcolor{blue}{($\uparrow$2.3)}}&63.9\textbf{\textcolor{blue}{($\uparrow$2.5)}}\\
\bottomrule
\end{tabularx}
\caption{Main experimental results on video reasoning and general video understanding benchmarks. All models are built on Qwen2.5-VL-7B-Instruct~\cite{bai2025qwen2} and trained on the same Video-R1-260k dataset~\cite{feng2025video} for fair comparison. 
$\textbf{↑x}$ denotes the performance improvement over Video-R1-7B (same frame setting).}
\label{tab:main_results}
\end{table*}

\subsection{Robust Optimization for Video Reasoning}
Training Video-LLMs with RL is notoriously unstable due to the sparsity of valid reasoning paths in complex video tasks.
We adapt two mechanisms to stabilize our 2D curriculum.

\subsubsection{Dynamic Sparse KL}
\label{sec:dynamic KL}
In high-difficulty regions of our grid (e.g., $B_{K,K}$), the model may initially fail to generate any correct responses, leading to reward collapse (all rewards are 0).
Standard RL objectives would force the model to minimize KL divergence with the reference model, causing Policy Degradation.
To counter this, we implement a Dynamic Sparse KL mechanism.
We dynamically monitor the reward variance $\sigma_r$ within each group.
If $\sigma_r = 0$ (indicating all samples are incorrect or correct), we detach the KL penalty term.
This allows the model to explore freely without being penalized for deviating from a reference model that likely also fails on these hard video samples.

\subsubsection{Structured Revisiting}
To address the risk of catastrophic forgetting common in course learning,
we propose a \textbf{Structured Revisiting} mechanism.
While random replay is a common method for mitigating forgetting,
we argue that this indiscriminate sampling approach is not optimal for post-video training.
Therefore, VideoCuRL does not uniformly sample historical data but instead selectively revisits orthogonal basis buckets from mastered lessons.
Specifically, this strategy prioritizes perceptual basis buckets ($B_{K,0}$,
high visual complexity and low text difficulty) to maintain the model's spatiotemporal tracking capabilities, while revisiting inference basis buckets ($B_{0,K}$,
low visual complexity and high text difficulty) to solidify logical deduction skills.
By prioritizing these basis buckets during replay, the model can efficiently and evenly retain and enhance its capabilities in both perception and reasoning dimensions with minimal computational overhead.

\section{Experiments}
\subsection{Main Results}
We present the comprehensive evaluation results in Table~\ref{tab:main_results}, comparing VideoCuRL-7B against the base model Qwen2.5-VL-7B-Instruct~\cite{bai2025qwen2} and the state-of-the-art RL baseline Video-R1-7B~\cite{feng2025video}.
Both baselines and our model are trained on the Video-R1-260k dataset to ensure a fair comparison.
The results demonstrate that VideoCuRL consistently outperforms both the SFT baseline and the standard RL approach across reasoning-intensive and general video understanding benchmarks.

\textbf{Superiority in Complex Reasoning}.
The core advantage of our orthogonal difficulty decomposition is most evident in reasoning-heavy tasks. On VSI-Bench~\cite{yang2025thinking}, which demands intricate causal deduction, VideoCuRL-7B achieves 37.3\% (16 frames) and 39.6\% (64 frames),
surpassing Video-R1-7B by margins of 2.7\% and 2.5\% respectively.
Similarly, on VideoMMMU~\cite{hu2025video} and MMVU~\cite{zhao2025mmvu},
our method consistently leads the leaderboard.

\begin{table*}[h]
\centering
\resizebox{\textwidth}{!}{
\begin{tabular}{lcccccc}
\toprule
\multirow{2}{*}{Model Variant} & \multicolumn{3}{c}{Video Reasoning Benchmarks} & \multicolumn{3}{c}{General Video Benchmarks} \\
\cmidrule(lr){2-4} \cmidrule(lr){5-7}
& VSI-Bench (16f) & VideoMMMU (32f) & MMVU (32f) & MVBench (16f) & TempCompass (32f) & VideoMME (64f) \\
\midrule
VideoCuRL-7B (Full Model) & 37.3 & 53.8 & 65.8 & 65.4 & 75.8 & 63.9 \\
\midrule
\multicolumn{7}{l}{\textit{Ablation 1: Single Difficulty Axis}} \\
- Only $D_{visual}$ & 35.8 (-1.5) & 52.6 (-1.2) & 64.5 (-1.3) & 65.8 (+0.4) & 74.3 (-1.5) & 63.2 (-0.7) \\
- Only $D_{text}$ & 36.1 (-1.2) & 52.1 (-1.7) & 64.2 (-1.6) & 62.3 (-3.1) & 74.2 (-1.6) & 62.8 (-1.1) \\
\midrule
\multicolumn{7}{l}{\textit{Ablation 2: Core Component Removal}} \\
- W/O Structured Revisiting & 35.7 (-1.6) & 51.9 (-1.9) & 63.7 (-2.1) & 62.8 (-2.6) & 73.5 (-2.3) & 62.1 (-1.8) \\
- W/O Dynamic Sparse KL & 34.9 (-2.4) & 51.7 (-2.1) & 63.9 (-1.9) & 63.5 (-1.9) & 73.1 (-2.7) & 62.2 (-1.7) \\
\midrule
\multicolumn{7}{l}{\textit{Ablation 3: Cognitive Metric Replacement}} \\
- Raw Perplexity ($PPL(A|Q)$) & 35.5 (-1.8) & 52.3 (-1.5) & 64.3 (-1.5) & 64.2 (-1.2) & 74.7 (-1.1) & 62.9 (-1.0) \\
\bottomrule
\end{tabular}
}
\caption{Ablation study results. All variants are based on VideoCuRL-7B. Numbers in parentheses denote performance change (↓) or (↑) compared to the full model. "16f/32f/64f" indicate frame settings.}
\label{tab:ablation}
\end{table*}

\textbf{Robustness in General Perception}.
A common pitfall in RL post-training is "Catastrophic Forgetting" of basic perception skills while chasing higher reasoning rewards.
However, VideoCuRL mitigates this effectively.
On MVBench~\cite{li2024mvbench},
a comprehensive perception benchmark,
our model maintains a clear lead, achieving 65.4\% accuracy at 16 frames compared to 62.7\% for Video-R1 and 57.4\% for the base model.
Notably, on TempCompass~\cite{liu2024tempcompass},
VideoCuRL achieves 75.0\% (16 frames) and 75.8\% (32 frames),
outperforming Video-R1 by significant margins ($\uparrow 2.4\%$ and $\uparrow 2.6\%$).
This empirical evidence confirms that our Structured Revisiting strategy successfully preserves and even refines fine-grained visual tracking skills.

\textbf{Comparison with Scalar and Generation-based Curricula.}
As shown in Table~\ref{tab:main_results},
Length-based Curriculum outperforms the random baseline Video-R1,
confirming that organizing data from easy to hard is generally beneficial.
However, it significantly lags behind VideoCuRL.
Furthermore, Generation-based Curriculum achieves competitive performance,
slightly behind VideoCuRL.
However, this comes at a prohibitive cost:
it requires several full inference passes over full dataset before training (Full implementation details are shown in Appendix \ref{sec:generation-based}),
consuming hundreds of GPU hours.
In contrast, VideoCuRL achieves superior performance using only training-free proxies ($D_{visual}$ and $D_{text}$) that are computed offline on CPUs. This demonstrates that our orthogonal decomposition effectively captures the true "hardness" of samples with dramatically higher efficiency.

\subsection{Experimental Setup}

\textbf{Implementation Details.}
We use Qwen2.5-VL-7B~\cite{bai2025qwen2} as our base model.
The 2D Curriculum Grid is set to $4\times4$ ($K=4$).
We provide a detailed sensitivity analysis regarding the choice of $K$ in Appendix \ref{sec:grid size},
demonstrating that $K=4$ strikes an optimal balance between curriculum granularity and training stability.
All models are trained using GRPO with a KL coefficient $\beta=0.04$.
Crucially, our proxy metric calculation is performed offline on CPUs,
incurring zero GPU inference cost for difficulty estimation.
For each pair of adjacent frames $(F_{t}, F_{t+1})$, we compute the dense optical flow field $V_{t}$ using the Gunnar Farneback algorithm~\cite{farneback2003two} implementation in OpenCV.
We use a pyramid scale of 0.5, 3 levels, and a window size of 15 to balance CPU efficiency and motion sensitivity.
We compute the entropy on grayscale difference maps.
Empirical analysis shows that luminance changes dominate temporal information density in video,
enabling efficient CPU-based computation without sacrificing the metric's discriminative power.

\textbf{Training Data \& Benchmarks}.
To strictly evaluate the efficacy of our curriculum strategy,
we utilize the exact same data split as the baseline Video-R1,
ensuring no data leakage.
For evaluation, we employ a comprehensive suite of benchmarks categorized into two groups:
\begin{figure*}
    \centering
    \includegraphics[width=\linewidth]{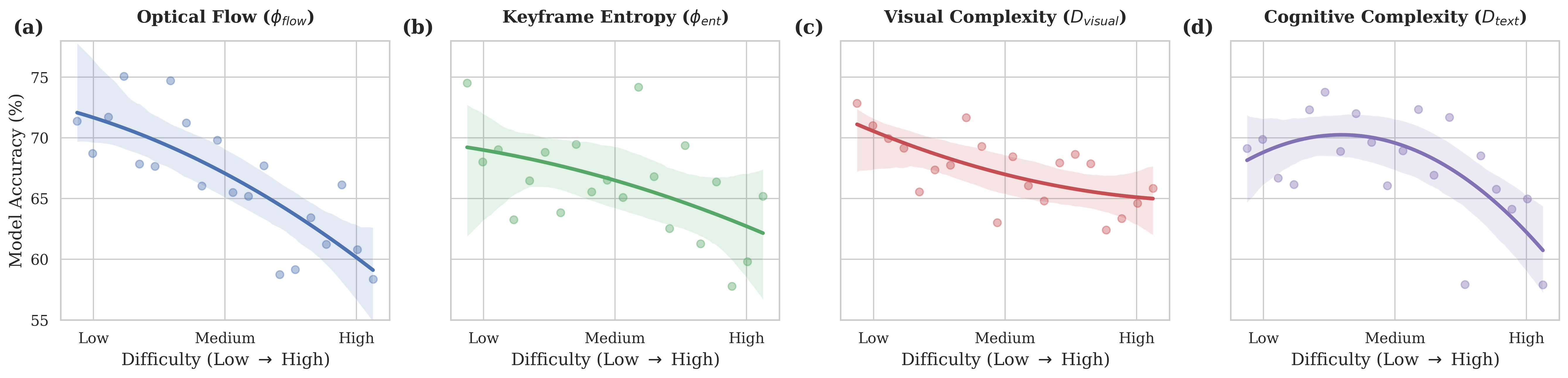}
    \caption{
    \textbf{Quantitative Cross-Validation of Difficulty Proxies.}
    We analyze the correlation between our training-free difficulty metrics and model accuracy on the training set.
    (a-b) Individual visual proxies ($\phi_{flow}$, $\phi_{ent}$) exhibit high variance, indicating that physical features alone are noisy predictors of difficulty. (c) The fused Visual Complexity ($D_{visual}$) smooths these fluctuations, providing a robust estimator of perceptual load. (d) Cognitive Complexity ($D_{text}$) reveals the nonlinear relationship between logical challenge and model performance.}
    \label{fig:cross valid}
\end{figure*}

Video Reasoning Benchmarks:
We use VSI-Bench~\cite{yang2025thinking}, VideoMMMU~\cite{hu2025video}, and MMVU~\cite{zhao2025mmvu} to assess the model's capability in causal deduction and complex problem-solving.

General Video Benchmarks:
We use MVBench~\cite{li2024mvbench} and TempCompass~\cite{liu2024tempcompass} to evaluate fundamental perception skills (e.g., object tracking, action recognition), and VideoMME~\cite{fu2025video} for long-context understanding.

We use LLaVA-onevision \cite{li2024llava}, LLaVA-video \cite{zhang2024video},
VILA \cite{lin2024vila},
Kangaroo \cite{liu2024kangaroo},
DeepVideo-R1 \cite{park2025deepvideo} for comparison.

\subsection{Ablation Studies}
To evaluate the contribution of each individual component in VideoCuRL, we conduct comprehensive ablation studies on both reasoning-heavy and perception-oriented benchmarks. The results are summarized in Table~\ref{tab:ablation}.

\subsubsection{Impact of Orthogonal Difficulty Decomposition}
We first investigate the necessity of disentangling visual and cognitive complexities by training two variants: one using only the Visual Complexity axis ($D_{visual}$) and another using only the Cognitive Reasoning axis ($D_{text}$). 

\textbf{Visual-only ($D_{visual}$)}:
While this variant performs relatively well on perception tasks like MVBench (+0.4\% over full model),
its performance on reasoning benchmarks (e.g., VSI-Bench) drops significantly by 1.5\%.
This confirms that perception-based scheduling alone is insufficient for mastering complex logical deduction.

\textbf{Cognitive-only ($D_{text}$)}:
Removing the visual dimension leads to a 3.1\% decline in MVBench and a 1.6\% drop in TempCompass.
This suggests that without a visual curriculum,
models suffer from "Alignment Tax,"
where they struggle to ground reasoning in complex temporal cues. 

\subsubsection{Effectiveness of Robust Optimization Mechanisms}
We further examine our two stabilization strategies: Structured Revisiting and Dynamic Sparse KL.

\textbf{W/O Structured Revisiting}:
Removing this mechanism leads to a drop across all benchmarks,
with the most notable impact on MVBench (-2.6\%) and TempCompass (-2.3\%).
This validates our hypothesis that random replay dilutes critical "basis" samples,
leading to catastrophic forgetting of fundamental perception skills.

\textbf{W/O Dynamic Sparse KL}:
This variant shows the sharpest decline in high-reasoning tasks, such as VSI-Bench (-2.4\%) and VideoMMMU (-2.1\%).
Without detaching the KL penalty in high-difficulty zones,
the model suffers from reward collapse and policy degradation,
as it is penalized for exploring beyond a failing reference model.

\subsubsection{Cognitive Metric: Calibrated Surprisal vs. Raw Perplexity}
Finally, we replace our Calibrated Surprisal (PMI-based) with standard Raw Perplexity (PPL).
The PPL-based variant yields inferior results across the board, particularly on VSI-Bench (-1.8\%) and VideoMMMU (-1.5\%).
This demonstrates that raw NLL is biased toward linguistic frequency,
whereas our calibrated metric more accurately captures the true logical information gain required for spatiotemporal reasoning.

\subsection{Empirical Validation of Difficulty Proxies}
\label{sec:cross difficulty}

To evaluate the reliability of our difficulty proxies,
we analyze the correlation between difficulty scores and model accuracy.
As shown in Figure \ref{fig:cross valid}(a-b),
raw physical metrics like Optical Flow $(\phi_{flow})$ and Keyframe Entropy $(\phi_{ent})$ exhibit high variance,
suggesting that individual motion features are insufficient for predicting model failure.
By fusing them into Visual Complexity $D_{visual}$,
we observe a stabilized inverse trend.
For Cognitive Complexity $D_{text}$,
the model accuracy exhibits a general downward trajectory,
although local fluctuations persist due to the stochastic nature of autoregressive generation and potential visual-linguistic misalignment.
The results of cross-validation confirms that our proxies effectively capture the statistical "hardness" of samples,
providing a robust signal for curriculum scheduling.
Detailed experimental setups are provided in Appendix \ref{sec:cross}.

\section{Conclusion}
In this paper, we propose VideoCuRL,
a novel Video Curriculum Reinforcement Learning framework.
By employing efficient difficulty proxies, VideoCuRL constructs a 2D curriculum grid navigated by a competence-aware Diagonal Wavefront strategy.
Extensive evaluations demonstrate that VideoCuRL consistently outperforms strong RL baselines and generation-based curricula without additional training or inference overhead.
Through Dynamic Sparse KL and Structured Revisiting, we also ensure training stability and mitigate catastrophic forgetting of fundamental perceptual skills.
Our results underscore that for complex multimodal reasoning,
strategic organization of data is as pivotal as the scale of the data itself.

\section*{Limitations}
Our framework prioritizes computational scalability by utilizing efficient, training-free difficulty proxies, a design choice that entails a deliberate trade-off between semantic granularity and processing efficiency. While our physical proxies effectively capture dynamic complexity without the prohibitive costs of generation-based oracles, they may under-represent the difficulty of visually subtle yet semantically dense scenes, such as micro-expressions. Similarly, the structural abstraction of difficulty into a discrete grid, while essential for orthogonal separation, introduces coarse-grained boundaries comp  ared to a theoretical continuous difficulty manifold, leaving room for future refinement in continuous curriculum modeling.

Additionally, as a post-training refinement framework, VideoCuRL operates under the standard cold-start assumption inherent to Reinforcement Learning. The methodology presupposes that the base SFT model possesses a foundational level of instruction-following capability to explore valid reasoning paths and initiate the wavefront expansion. Consequently, our approach is positioned as a competence amplifier for aligned models rather than a solution for learning multimodal representations completely

\bibliography{custom}
\appendix
\clearpage

\section{Preliminaries: Group Relative Policy Optimization (GRPO)}
\label{app:grpo_background}

In this work, we employ Group Relative Policy Optimization (GRPO) as our core reinforcement learning algorithm. While Proximal Policy Optimization (PPO)~\cite{schulman2017proximal} is a standard choice for aligning Large Language Models, it typically requires maintaining a value function (Critic) model roughly the same size as the policy model. In the context of Video-LLMs, where visual encoders and long-context histories already impose significant GPU memory overhead, maintaining a separate Critic is often computationally prohibitive.

GRPO circumvents the need for a value network by estimating the baseline directly from a group of sampled outputs. Formally, the training process is defined as follows:

\paragraph{Group Sampling.} 
Given a batch of inputs consisting of video clips $V$ and textual queries $Q$, we sample a group of $G$ outputs $\{A_1, A_2, \dots, A_G\}$ from the current policy $\pi_{\theta_{old}}$. This allows the model to explore diverse reasoning paths for the same visual-linguistic context.

\paragraph{Advantage Estimation.} 
Instead of relying on a learned value function $V(S)$ to compute the advantage, GRPO utilizes the group statistics.
For each output $A_i$ in the group,
we compute a specific reward $r_i$ (based on correctness or rule adherence).
The advantage $Adv_i$ is then calculated by normalizing the rewards within the group:
\begin{equation}
    Adv_i = \frac{r_i - \text{mean}(\mathbf{r})}{\text{std}(\mathbf{r}) + \epsilon}
\end{equation}
where $\mathbf{r} = \{r_1, \dots, r_G\}$ is the set of rewards for the group, and $\epsilon$ is a small constant for numerical stability.
This effectively uses the average performance of the group as the baseline, encouraging the model to reinforce outputs that perform better than the expected average for that specific video-query pair.

\paragraph{Objective Function.} 
The final optimization objective maximizes the surrogate loss while constraining the policy update via KL divergence, similar to PPO. The GRPO objective is formulated as:
\begin{equation}
\begin{gathered}
    \mathcal{J}_{GRPO}(\theta) = \mathbb{E}_{Q \sim \mathcal{D}, \{A_i\}_{i=1}^G \sim \pi_{\theta_{old}}} \bigg[ \\
    \frac{1}{G}\sum_{i=1}^G \bigg(\min \Big( \rho_i Adv_i, \text{clip}(\rho_i, 1{\pm}\epsilon) Adv_i \Big) \\
    - \beta D_{KL}(\pi_\theta || \pi_{ref}) \bigg) \bigg]
\end{gathered}
\label{eq:grpo}
\end{equation}
where $\rho_i = \frac{\pi_{\theta}(A_i|V, Q)}{\pi_{\theta_{old}}(A_i|V, Q)}$ is the probability ratio, clip$(\cdot)$ limits the update step size, and $D_{KL}$ represents the KL divergence between the current policy $\pi_\theta$ and the reference SFT model $\pi_{ref}$ to prevent reward hacking.

In our VideoCuRL framework, we further modify this objective by introducing \textit{Dynamic Sparse KL} (as detailed in Section~\ref{sec:dynamic KL}) to handle the reward collapse issues inherent in high-difficulty curriculum stages.

\section{Implementation of Generation-based Curriculum Baseline}
\label{sec:generation-based}

To further validate the efficiency and effectiveness of VideoCuRL,
we introduced Generation-based Curriculum as a strong benchmark in our experiments.
This approach represents the "model-as-teacher" strategy commonly used in the industry when explicit physical metrics are lacking.
Its core logic is to pre-evaluate the training dataset $D$ (i.e., Video-R1-260k) using a pre-trained LMMs,
quantifying the difficulty based on the model's actual response to each sample.

Specifically, this method requires a pre-trained Video-LLM to perform complete inference on 260k video-instruction pairs. The difficulty score is determined by the accuracy generated by the model. By calculating the average accuracy of the model generating multiple benchmark answers $A$, we linearly divide the dataset into multiple stages from easy to difficult and train a single-dimensional linear course using the same learning rate and number of training steps as VideoCuRL.
\begin{table*}[ht]
\centering
\begin{tabular}{lcccc}
\toprule
\textbf{Grid Size ($K$)} & \textbf{VSI-Bench (16f)} & \textbf{MVBench (16f)} & \textbf{VideoMME (64f)} & \textbf{VideoMMMU (64f)} \\ \midrule
$K=3$ ($3 \times 3$)    & 35.8\%                   & 63.2\%                 & 62.4\%                  & 53.1\%                 \\
\textbf{$K=4$ ($4 \times 4$)} & \textbf{37.3\%}          & \textbf{65.4\%}        & \textbf{63.9\%}         & \textbf{54.5\%}      \\
$K=5$ ($5 \times 5$)    & 37.5\%                   & 65.7\%                 & 64.1\%                  & 54.3\%                \\ \bottomrule
\end{tabular}
\caption{Sensitivity analysis of the grid size $K$ on video reasoning and perception benchmarks. $K=4$ provides the optimal balance between granularity and stability.}
\label{tab:k_sensitivity}
\end{table*}

However, this approach is limited by its extremely high pre-computational overhead. Due to the need for multiple autoregressive decodings of massive amounts of video data, this process consumes hundreds of GPU hours in the preprocessing stage, which is often unsustainable when dealing with large-scale video datasets. In contrast, VideoCuRL achieves equivalent accuracy in difficulty assessment on a CPU using lightweight visual proxy metrics such as optical flow and entropy, along with text-based surprise calibration, without any GPU inference cost.

Experimental results (see Table~\ref{tab:main_results} in the main text) show that although the generative approach achieved a VSI-Bench accuracy of 36.5\%, it was still slightly inferior to VideoCuRL-7B in cognitive reasoning and complex perceptual tasks. This indicates that relying solely on scalarized model feedback is insufficient to capture the orthogonality between "perceptual load" and "cognitive depth" in video understanding, while VideoCuRL's two-dimensional decoupled architecture significantly improves the cost-effectiveness and robustness of training while maintaining high performance.

\section{Implementation Details of Quantitative Cross-Validation}
\label{sec:cross}

To ensure transparency, we detail the methodology and the statistical characteristics of the validation results presented in Section \ref{sec:cross difficulty}.

\textbf{Data Sampling and Inference Protocol}.
We randomly selected a validation subset of 1k video-instruction pairs from the Video-R1-260k dataset.
We used the base model (Qwen2.5-VL-7B-Instruct) to perform zero-shot inference. For each sample, the response was scored as 1 (correct) or 0 (incorrect) based on ground-truth verification.

\textbf{Binning Analysis and Statistical Observations}.
To visualize the correlation in Figure \ref{fig:cross valid}, we adopted a binning approach:

\textit{Horizontal Axis}:
We sorted samples by proxy scores ($\phi_{flow}$, $\phi_{ent}$, $D_{visual}$, and $D_{text}$) and applied Quantile Binning to create 20 equal-sized intervals.

\textit{Vertical Axis}:
We calculated the mean accuracy within each bin.

\textit{Observations on Variance}: Unlike generation-based estimators that require multiple inference passes, our training-free proxies do not account for the model's internal state.
Consequently,
Figure \ref{fig:cross valid} reflects heteroscedasticity—the variance in model performance remains significant even in low-difficulty bins.
This suggests that while $D_{text}$ identifies samples requiring deep logical deduction, failure can still occur on "easy" samples due to unrelated factors such as visual occlusions or linguistic ambiguity.

\textbf{Interpretation of the Cognitive Proxy}.
The Calibrated Surprisal ($D_{text}$) measures the Information Gain required to reach the answer.
The observed correlation confirms that as the logical entropy between the question and answer grows,
the probability of the model maintaining a correct reasoning chain diminishes.
By acknowledging the inherent noise in these proxies, VideoCuRL utilizes them as probabilistic guides rather than deterministic labels,
ensuring a robust curriculum that is resilient to individual sample outliers.

\section{Sensitivity Analysis of Grid Size $K$}
\label{sec:grid size}
we investigate the impact of the grid resolution $K$ on the overall performance of VideoCuRL. The parameter $K$ determines the granularity of our orthogonal difficulty decomposition, where a $K \times K$ grid partitions the dataset into $K^2$ distinct buckets.

\textbf{Experimental Setup}.
We conducted experiments with $K \in \{3, 4, 5\}$ while keeping all other hyperparameters, such as the KL coefficient $\beta=0.04$ and total training steps, constant.
All variants were trained on the Video-R1-260k dataset using the Diagonal Wavefront scheduling strategy.
Performance was evaluated across three representative benchmarks: VSI-Bench (Reasoning), MVBench (Perception), and VideoMME (Long-context).

As shown in Table \ref{tab:k_sensitivity},
Increasing $K$ from 3 to 4 yields a significant performance gain, particularly in reasoning tasks (+1.5\% on VSI-Bench).
This suggests that a $3 \times 3$ grid is too coarse to effectively disentangle the complex interplay between visual perception load and cognitive depth.
While $K=5$ provides marginal improvements over $K=4$, the gains are statistically insignificant. However, a $5 \times 5$ grid increases the number of buckets from 16 to 25, leading to higher computational overhead in tracking Local Competence Scores (LCS) and managing the wavefront expansion.
We observed that $K=4$ provides the most stable training trajectory. With $K=5$, some high-difficulty buckets (e.g., $B_{4,4}$) contain fewer samples, occasionally leading to increased reward variance and triggering the Dynamic Sparse KL mechanism more frequently than necessary.

\textbf{Conclusion}.
Based on these findings,
we selected $K=4$ as the default configuration for VideoCuRL.
It strikes an optimal balance between curriculum precision and optimization stability,
ensuring robust spatiotemporal reasoning without excessive partitioning of the training data.

\end{document}